\documentclass{article} 
\usepackage{aditi}
\renewcommand{\ForumContactRow}{%
  \begingroup\small\raggedright
    \ifx\ForumEmail\empty\else
      {\color{ForumAccent}\faEnvelope[regular]~}\ %
      \href{mailto:\ForumEmail}{\textcolor{ForumContactText}{\texttt{\ForumEmail}}}\par
      \vspace{\ForumContactGap}%
    \fi
    
  \endgroup
}
\usepackage{microtype}
\usepackage{xspace}
\usepackage{hyperref}
\usepackage{url}
\usepackage{booktabs}
\usepackage{multirow}
\usepackage{multicol}
\usepackage{lineno}
\usepackage{style} 
\usepackage{array}
\usepackage{hyperref}
\usepackage{fontawesome5} 
\definecolor{skyblue}{RGB}{204,229,255}

\usepackage{algorithm}
\usepackage{algorithmic}

\usepackage{tabularx}
\newcolumntype{Y}{>{\centering\arraybackslash}X}

\usepackage[most]{tcolorbox}
\tcbset{
  lightbluebox/.style={
    colback=skyblue!70,        
    colframe=skyblue!75!black, 
    boxrule=1pt,
    arc=4pt,
    auto outer arc
  }
}

\definecolor{darkblue}{rgb}{0, 0, 0.5}
\hypersetup{colorlinks=true, citecolor=darkblue, linkcolor=darkblue, urlcolor=darkblue}

\newcommand{\method}{\textsc{ViewFusion}\xspace}

\setTitleruleGap{0.75pt}         

\title{ViewFusion: Structured Spatial Thinking Chains for Multi-View Reasoning}

\setauthors{Xingjian Tao$^{1}$ \authorsep Yiwei Wang$^{3}$\authorsep Yujun Cai$^{4}$  \authorsep Yifan Song$^{1}$   \authorsep Jing Tang$^{1,2}$\thanks{Corresponding Author: Jing Tang.}}
\setaffils{$^{1}$HKUST(GZ),\quad $^{2}$HKUST, \quad $^{3}$UC Merced, \quad $^{4}$University of Queensland}

\setemail{taoxj2001@outlook.com}
\setemails{\href{https://github.com/taoxj2001/ViewFusion}{\faGithub\ https://github.com/taoxj2001/ViewFusion}}

\usepackage{xcolor}
\usepackage{minitoc}
\usepackage{graphicx}
\definecolor{jcg}{RGB}{100,160,0}
\definecolor{sachin}{RGB}{0,0,150}
\definecolor{hqz}{RGB}{160,100,100}
\definecolor{gnz}{HTML}{64B5F6}
\usepackage[most]{tcolorbox} 
\tcbuselibrary{skins,breakable} 

\definecolor{myDarkGreen}{RGB}{50, 70, 70} 
\definecolor{myLightGray}{RGB}{240, 240, 240} 

\definecolor{titlebgcolor}{RGB}{70, 80, 100}
\definecolor{bodybgcolor}{RGB}{245, 245, 245}
\definecolor{bordercolor}{RGB}{120, 120, 120}
\definecolor{darkblue}{rgb}{0.0, 0.0, 0.55}   
\definecolor{darkgreen}{rgb}{0.0, 0.5, 0.0}   
\definecolor{darkred}{rgb}{0.6, 0.0, 0.0}     

\usepackage{url}
\usepackage{amsthm}
\usepackage{minitoc}
\usepackage{enumitem}
\usepackage{todonotes}
\usepackage{float}
\usepackage{listings}
\usepackage{caption}
\usepackage{cleveref}
\usepackage[table]{xcolor} 
\usepackage{graphicx}

\usepackage{booktabs}

\definecolor{myLightBlue}{RGB}{230, 240, 255} 

\usepackage[T1]{fontenc}

\newtcolorbox{responsebox}[2][]{
    breakable,
    enhanced,
    colback=white,             
    colframe=blue!50!black,    
    coltext=black,             
    coltitle=white,    
    fonttitle=\bfseries\rmfamily, 
    arc=3mm,                   
    boxrule=1pt,
    title=#2,
    #1
}

\definecolor{lightblue}{RGB}{235,243,252}

\definecolor{mybgcolor}{RGB}{235, 235, 250}
\definecolor{myGreen}{RGB}{240, 250, 240}

\newtcolorbox{takeawaybox}[1][]{
  enhanced,
  colback=mybgcolor, 
  colframe=black,    
  boxrule=0.5pt,     
  arc=3mm,           

  attach boxed title to top left={yshift=-0.25em, xshift=1em},
  fonttitle=\bfseries, 
  title={#1},          
  boxed title style={
    colback=black,     
    sharp corners,     
  },
}
\newtcolorbox{equationbox}[1]{
  colback=white,                
  colframe=gray!75!black,       
  boxrule=1pt,                  
  
  title=#1,                     
  attach boxed title to top left={yoffset=-2mm, xshift=2mm}, 
  
  colbacktitle=gray!75!black,   
  coltitle=white,               
  fonttitle=\bfseries\sffamily, 
  
  boxed title style={
    boxrule=0pt,                
    frame code={}               
  }
}

\begin{document}

\maketitle

\begin{abstract}
Multi-view spatial reasoning remains difficult for current vision-language models. Even when multiple viewpoints are available, models often underutilize cross-view relations and instead rely on single-image shortcuts, leading to fragile performance on viewpoint transformation and occlusion-sensitive cases. We present \method, a two-stage framework that explicitly separates cross-view spatial pre-alignment from question answering. In the first stage, the model performs deliberate spatial pre-thinking to infer viewpoint relations and spatial transformations across views, forming an intermediate workspace that goes beyond a simple re-description. In the second stage, the model conducts question-driven reasoning conditioned on this workspace to produce the final prediction. We train \method with synthetic reasoning supervision followed by reinforcement learning using GRPO, which improves answer correctness while stabilizing the intended two-stage generation behavior. On MMSI-Bench, \method improves accuracy by 5.3\% over Qwen3-VL-4B-Instruct, with the largest gains on examples that require genuine cross-view alignment.
\end{abstract}

\section{Introduction}




Recent advances in vision-language models have enabled machines to understand 
and reason about visual content alongside natural language. Models such as LLaVA~\citep{liu2023visual, liu2024improved, li2024llava}, Flamingo~\citep{alayrac2022flamingo}, Gemini~\citep{team2023gemini}, and Qwen-VL~\citep{bai2023qwen,wang2024qwen2,bai2025qwen2} enable more comprehensive reasoning across modalities, supporting tasks like visual question answering, image captioning, and document understanding where both language and vision are crucial. However, multi-view  spatial reasoning, the ability to align spatial information across different viewpoints and reason about 3D scene structure, remains a fundamental challenge that current models struggle to solve reliably.

The core difficulty lies in cross-view spatial alignment. When presented with multiple images of the same scene from different viewpoints, models must not only recognize objects and their attributes within each view, but also establish spatial correspondences across views: How has the camera moved? Which objects correspond under viewpoint change? How do occlusions evolve as perspective shifts? These cross-view relations are essential for answering questions that require genuine multi-view understanding, yet they remain largely implicit in current reasoning approaches.

 \begin{figure}[t!]
    \centering
        \includegraphics[width=1\linewidth]{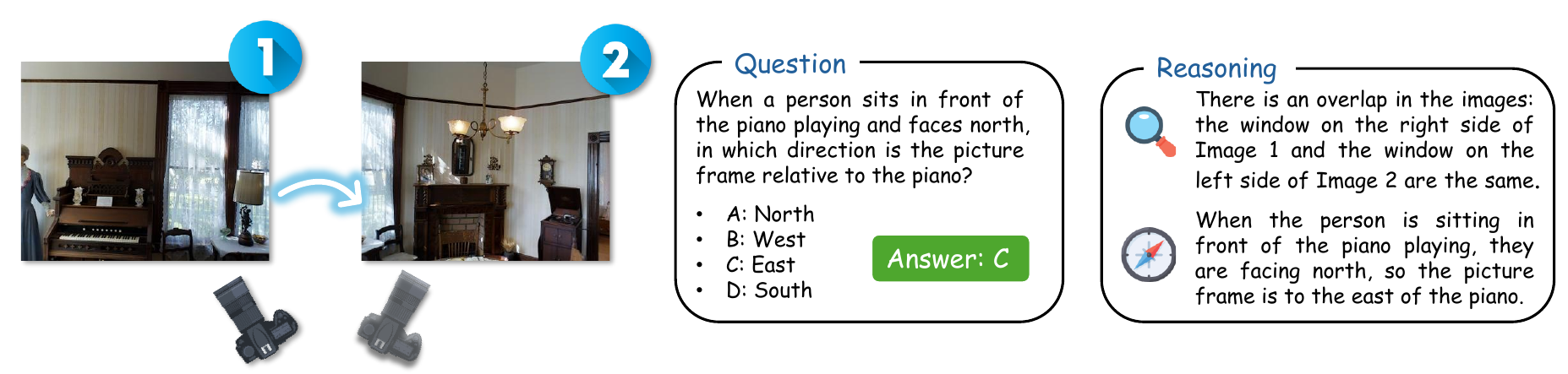}
        \caption{An example of multi-view spatial reasoning. Given two images captured from different viewpoints, the model must align shared visual cues across views to infer the viewpoint relationship and answer a direction-based question (e.g., locating the picture frame relative to the piano)}

        \label{fig:example}
        
\end{figure}

Consider a concrete example illustrated in~\Cref{fig:example}. Given two living-room images captured from different viewpoints, a question asks: “When a person sits in front of the piano playing and faces north, in which direction is the picture frame relative to the piano?” In the first view, the piano is seen beside a tall window and the picture frame is not clearly localized; in the second view, the scene is observed from a shifted angle where the window and wall decor provide an overlap across views and the picture frame becomes visible on the corresponding wall. To answer correctly, the model must align these shared cues to infer the relative viewpoint change and then map the picture frame’s position into the question’s north-referenced coordinate frame. Without proper cross-view alignment, the model may mis-associate landmarks across images or reason from a single view, resulting in an incorrect direction even when each individual image is well described.

In practice, many existing approaches~\citep{liu2023visual, tong2024cambrian, chen2024spatialvlm} still rely heavily on single-view cues and superficial correlations, which leads to brittle behavior and noticeable performance drops when multiple viewpoints must be aligned and complementary evidence integrated to resolve ambiguities. This limitation also persists when adopting reinforcement learning as a post-training strategy. While RL methods such as Group Relative Policy Optimization (GRPO) can improve task-level performance from model rollouts~\citep{guo2025deepseek} and often encourage more elaborate deliberation, they do not necessarily induce correct cross-view spatial alignment. In our preliminary study, models trained with vanilla GRPO frequently exhibit shortcut behaviors: they begin solving the question before integrating the full multi-view context, or rely predominantly on a single view while treating other images as incidental. Consequently, the resulting reasoning may appear detailed but remains grounded in an incomplete cross-view spatial model, and providing more views does not reliably translate into better multi-view reasoning. In some cases, additional views can even introduce noise that amplifies shortcut learning and destabilizes intermediate reasoning.

To address the persistent difficulty of establishing correct cross-view spatial relations, we propose \method, a simple but effective two-stage “think twice” paradigm for multi-view spatial reasoning. \method explicitly separates spatial pre-thinking from question answering. In the first stage, the model performs deliberate spatial pre-thinking to infer viewpoint relations and spatial transformations across views and organize them into an intermediate workspace. In the second stage, the model conducts question-driven reasoning conditioned on this workspace to produce the final answer. The core principle is to make cross-view alignment a deliberate first step, rather than an implicit byproduct of answering. We train \method by first performing supervised fine-tuning with synthesized reasoning traces that reflect this two-stage protocol, and then applying reinforcement learning with GRPO to further align the model toward correct answers while stabilizing consistent two-stage behavior and reliable generation.

Empirically, \method achieves strong results on MMSI-Bench~\citep{yang2025mmsi}, improving accuracy by 5.3\% over Qwen3-VL-4B-Instruct, with particularly large gains on examples that require genuine cross-view alignment. We compare \method with several popular spatial reasoning baselines and observe consistent improvements across settings. Our ablation studies further validate the reliability of each component in the proposed design. Notably, \method also outperforms Qwen3-VL-4B-Thinking, suggesting that explicitly enforcing a two-stage pre-thinking protocol yields benefits beyond simply encouraging longer deliberation. Our contributions can be summarized as follows:

\begin{tcolorbox}[
  enhanced, breakable,
  colframe=black!12, boxrule=0.35pt, arc=1mm,
  title={\textbf{Summary of Our Main Contribution}},
  coltitle=black, fonttitle=\sffamily\bfseries,
  colbacktitle=blue!15!white,  
  colback=blue!5!white,      
  boxed title style={
    sharp corners, boxrule=0pt,
    top=3pt, bottom=3pt, left=4mm, right=4mm,
    borderline={0.5pt}{0pt}{black!10}       
  },
  attach boxed title to top left={xshift=4mm,yshift*=-1.2mm},
  boxsep=1.5mm, top=1.5mm, bottom=1.5mm, left=4mm, right=4mm,
  before skip=10pt, after skip=10pt
]
\begin{enumerate}[topsep=0pt,leftmargin=10pt]\setlength{\itemsep}{0pt}
\item We diagnose a key failure mode in current multi-view spatial reasoning with MLLMs, including RL-trained models: they often fail to align cross-view spatial information and instead rely on shortcut behaviors that underuse the available viewpoints.
\item We introduce \method, a two-stage “think twice” paradigm that explicitly separates cross-view spatial pre-thinking from question solving, together with a training recipe that combines synthesized reasoning supervision and GRPO-based reinforcement learning to stabilize this behavior. 
\item We provide comprehensive experimental evidence on MMSI-Bench, including comparisons with strong and widely used baselines (notably Qwen3-VL-4B-Instruct and Qwen3-VL-4B-Thinking) and ablations that verify the contribution of each component.
\end{enumerate}
\end{tcolorbox}

 \begin{figure}[t!]
    \centering
 
        \includegraphics[width=1\linewidth]{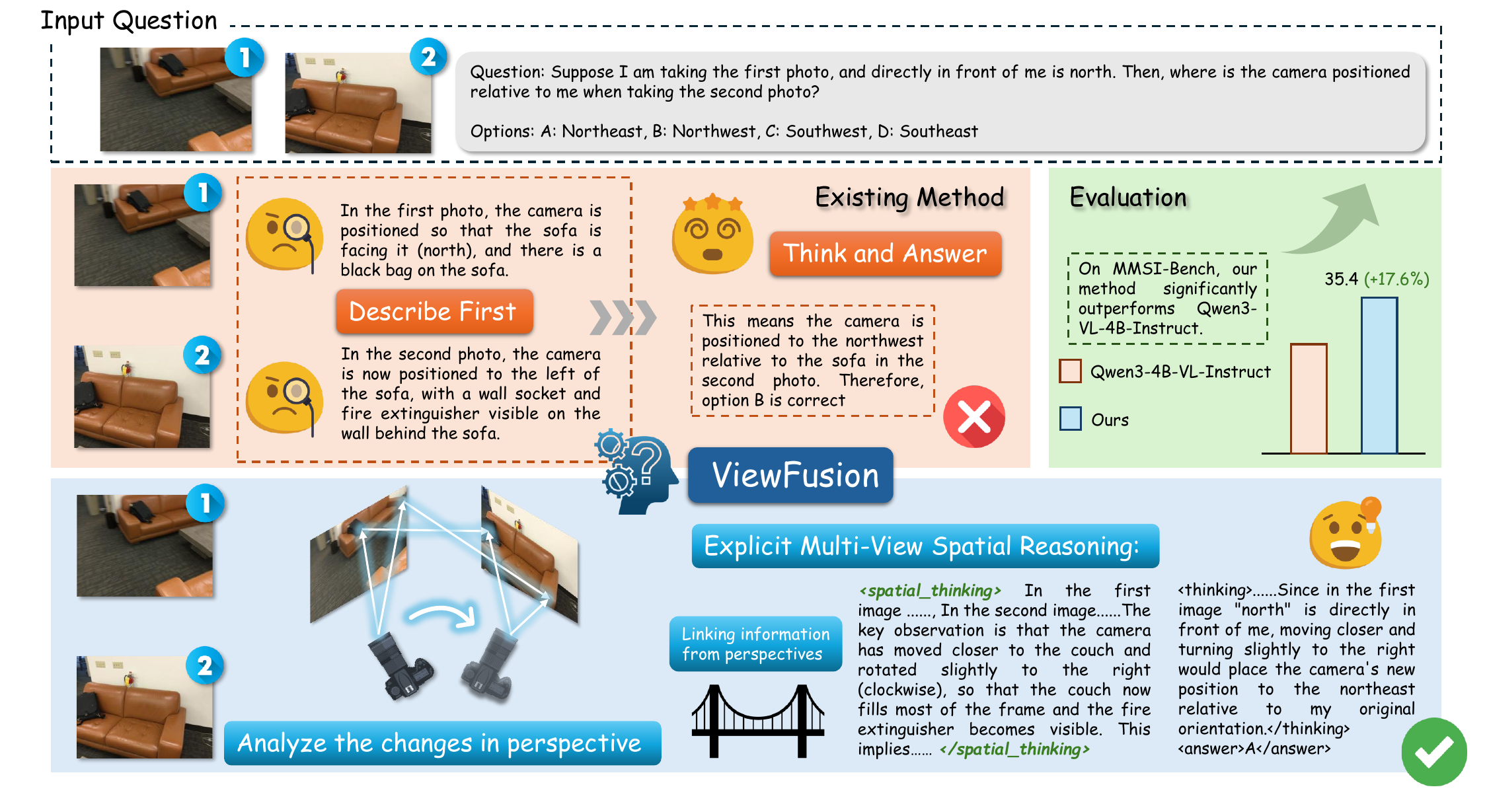}
      
        \caption{Overview of \method for multi-view spatial reasoning. Given a multi-view question (left), existing “describe-first” or direct “think-and-answer” paradigms often produce view-local descriptions and then shortcut to answering without establishing correct cross-view spatial relations, leading to errors (top). \method instead performs explicit multi-view spatial pre-thinking to link perspectives and infer viewpoint transformations across images before question solving (bottom), yielding more reliable reasoning and correct predictions.}
    
        \label{fig:pkpo_entropy}
        
\end{figure}
 \section{Related Work}\label{sec:related work}

\subsection{Reinforcement Learning for MultiModal Large Language Models Reasoning}
Reinforcement learning (RL) and preference optimization have become increasingly popular for improving the reasoning quality and behavioral alignment of multimodal large language models (MLLMs)~\citep{sun2024aligning, yu2024rlhf, yu2024rlaif, xie2024v,huang2025vision}. Beyond outcome-level supervision, recent reasoning-oriented RL methods aim to provide richer training signals that shape the structure of multimodal reasoning. For example, Insight-V~\citep{dong2025insight} leverages a multi-agent setup to select and learn from self-generated reasoning trajectories, while R1-VL~\citep{zhang2025r1} introduces step-wise GRPO with dense rule-based rewards to improve multimodal reasoning paths. This GRPO-style training has also been extended to video reasoning, including Video-R1~\citep{feng2025video}, Video-RTS~\citep{wang2025video}, and Video-STR~\citep{wang2025video2}. In parallel, several works encourage models to “observe first” by introducing explicit context descriptions or observation stages, such as HumanOmniV2~\citep{yang2025humanomniv2}, Visionary-R1~\citep{xia2025visionary}, and Observe-R1~\citep{xia2025visionary}. However, these observation steps are typically realized as descriptive summaries of the visual input, and do not explicitly induce the model to reason about relationships across multiple views. Overall, existing RL-based approaches can encourage stronger deliberation and better alignment in MLLMs, yet designing rewards and training protocols that explicitly promote cross-view spatial consistency remains an open challenge for multi-view reasoning.

\subsection{Spatial reasoning with MLLMs}
Spatial reasoning has emerged as a key frontier for MLLMs, aiming to move beyond object recognition toward understanding relative position, orientation, viewpoint transformation, and occlusion-aware relations in 3D scenes. A growing body of work seeks to strengthen spatial reasoning in MLLMs by improving grounding and spatial representations, for example via spatially aware instruction tuning and curated supervision~\citep{liu2023visual2, tong2024cambrian, chen2024spatialvlm,yu2025far,gholami2025spatial,wu2025spatialscore,zhao2025spacemind, batra2025spatialthinker, wang2025visioncube, fan2025vlm, li2024topviewrs}. More recently, Visual Spatial Tuning~\citep{yang2025visual} trains vision-language models with large-scale spatial perception and reasoning data, producing notably stronger spatial reasoning performance and improved generalization across spatial benchmarks~\citep{yang2025visual}. 

To systematically evaluate these advances under multi-view inputs, several benchmarks have been proposed~\citep{zhang2025sphere, lee2025spatialmosaic}. MMSI-Bench~\citep{yang2025mmsi} focuses on multi-view, multi-image spatial intelligence and includes problems that require aligning evidence across views rather than solving from a single snapshot. MindCube~\citep{yin2025spatial} probes whether models can construct and manipulate a coherent “mental model” of the scene from limited observations, emphasizing perspective taking and spatial consistency under incomplete information. ViewSpatial~\citep{li2025viewspatial} further stresses viewpoint-dependent spatial localization and cross-view reference frames, revealing substantial generalization gaps when camera viewpoints shift. Collectively, these benchmarks provide increasingly fine-grained diagnostics for multi-view spatial understanding and consistently highlight a key bottleneck: strong single-view perception does not automatically translate into reliable cross-view alignment, leaving significant room for methods that explicitly model and enforce spatial consistency across views.

 \section{\method}

\subsection{Limitations of Reasoning Models under Multi-View Inputs}
Existing reasoning paradigms for multi-view inputs often fall short because they do not explicitly infer the spatial relationships between images captured from different viewpoints. Instead of establishing cross-view consistency (e.g., how the camera moves, which objects correspond across views, and how occlusions change), the model frequently treats each image as an independent evidence source and proceeds directly to question answering. This ``late fusion'' behavior implicitly assumes that the relevant information is already visible and interpretable within a single view, so multi-view inputs are used as weak auxiliary context rather than as complementary observations that must be jointly aligned.

Even when an intermediate ``observation'' or description step is introduced, it is typically view-local and descriptive, summarizing salient entities within each frame without reasoning about how these entities transform across viewpoints. Such descriptions often omit the most informative cues for multi-view tasks, including which objects disappear due to occlusion versus leaving the scene, how the relative ordering of background landmarks changes under camera motion, and how scale or perspective distortions affect apparent positions. 

As a result, key spatial cues that only emerge through cross-view alignment---such as viewpoint-dependent visibility, relative layout changes, and object re-identification under occlusion---are easily missed. This leads to reasoning traces that appear coherent but are grounded in an incomplete spatial model, making the final prediction brittle when the task requires genuine multi-view integration. In our error analysis, these failures frequently manifest as plausible but inconsistent narratives (e.g., treating two views as different locations, confusing left/right after a turn, or matching objects to incorrect counterparts) that cannot be corrected by additional textual deliberation alone.

Motivated by these limitations, we argue that effective multi-view reasoning requires an explicit pre-alignment step that prioritizes cross-view spatial consistency before any question-driven inference. Rather than asking the model to ``solve while observing'', we enforce a ``pre-think then answer'' protocol in which cross-view alignment is a deliberate first step. Specifically, the model first infers viewpoint relations and spatial transformations across images to form a consistent intermediate workspace, then performs task-specific reasoning conditioned on this workspace. This decomposition targets the shortcut behaviors observed in existing paradigms by (i) forcing the model to reconcile multi-view evidence before committing to an answer, and (ii) making alignment errors more visible and thus easier to constrain during training. As a result, the model becomes more robust on cases where the answer is only recoverable through genuine multi-view integration, such as questions requiring viewpoint transformation, occlusion-aware reasoning, and cross-view correspondence.

\subsection{Training Data Preparation}
Our training data is sampled from two multi-view datasets, VST-500K~\citep{yang2025visual} and the MindCube-Trainset~\citep{yin2025spatial}, and is organized into two splits corresponding to the two-stage training pipeline: supervised fine-tuning (SFT) and reinforcement learning (RL).

\paragraph{SFT data (18K).}
We construct an SFT set containing 18K multi-view instances. For each instance, we rewrite the original rationale into a structured reasoning trace using Qwen3-32B-Instruct. Each trace contains three parts: \texttt{<spatial\_thinking>}, \texttt{<thinking>}, and \texttt{<answer>}. The \texttt{<spatial\_thinking>} part is designed to elicit an explicit pre-thinking process that focuses on establishing spatial relations across views (e.g., viewpoint change and cross-view consistency) before proceeding to question-driven reasoning in \texttt{<thinking>} and producing the final prediction in \texttt{<answer>}. To ensure training stability and format controllability, we apply strict filtering rules and remove samples that violate the required structure (e.g., missing fields, incorrect ordering, unclosed tags, or malformed outputs). This yields a clean SFT corpus with consistent two-stage reasoning behavior.

\paragraph{RL data (16K).}
We additionally construct an RL set of 16K instances from the same data sources. Unlike the SFT set, the RL split does not include rewritten reasoning traces; it only retains the multi-view input and the supervision necessary for computing outcome-based rewards (e.g., final answer correctness). This separation allows RL to further align the policy toward task success while avoiding overfitting to specific rationales, and enables us to explicitly study how RL interacts with the proposed two-stage reasoning protocol.

\subsection{Training Strategy}
\label{sec:training}

\subsubsection{Preliminary: SFT and GRPO}
\label{sec:prelim_sft_grpo}

We briefly review supervised fine-tuning (SFT) and Group Relative Policy Optimization (GRPO), which serve as the two training stages of our framework.

\paragraph{Supervised Fine-Tuning (SFT).}
Given a dataset of multimodal instruction-response pairs
$\mathcal{D}=\{(x_i, y_i)\}_{i=1}^{N}$,
where $x_i$ denotes the input (e.g., text with one or multiple images) and $y_i$ the target output, SFT trains a conditional generative policy $\pi_{\theta}(y \mid x)$ by maximizing the log-likelihood:
\begin{equation}
\label{eq:sft}
\mathcal{L}_{\mathrm{SFT}}(\theta)
= - \frac{1}{N} \sum_{i=1}^{N} \log \pi_{\theta}(y_i \mid x_i).
\end{equation}

\paragraph{Group Relative Policy Optimization (GRPO).}
Reinforcement learning further optimizes the policy using a reward function $r(x,y)$ defined on model outputs.
For each input $x$, GRPO samples a group of $K$ candidate outputs
$\{y^{(k)}\}_{k=1}^{K} \sim \pi_{\theta}(\cdot \mid x)$,
and computes a group-wise baseline to reduce variance:
\begin{equation}
\label{eq:grpo_adv}
A^{(k)} = r(x, y^{(k)}) - \frac{1}{K}\sum_{j=1}^{K} r(x, y^{(j)}).
\end{equation}
GRPO then updates $\theta$ by maximizing a PPO-style clipped objective with a KL regularizer that anchors the policy to a reference model $\pi_{\mathrm{ref}}$:
\begin{equation}
\label{eq:grpo_obj}
\begin{aligned}
\mathcal{L}_{\mathrm{GRPO}}(\theta)
=
&\ \mathbb{E}_{x \sim \mathcal{D}}
\left[
\frac{1}{K}\sum_{k=1}^{K}
\min\!\left(
\rho^{(k)} A^{(k)},
\mathrm{clip}\!\left(\rho^{(k)}, 1-\epsilon, 1+\epsilon\right) A^{(k)}
\right)
\right] \\
&\ - \beta \, \mathbb{E}_{x}\!\left[
D_{\mathrm{KL}}\!\left(\pi_{\theta}(\cdot \mid x)\,\|\,\pi_{\mathrm{ref}}(\cdot \mid x)\right)
\right],
\end{aligned}
\end{equation}
where $\rho^{(k)} = \frac{\pi_{\theta}(y^{(k)} \mid x)}{\pi_{\theta_{\mathrm{old}}}(y^{(k)} \mid x)}$,
$\epsilon$ is the clipping threshold, and $\beta$ controls the KL strength.

\subsubsection{Two-Stage Optimization}
\label{sec:two_stage_training}

We train \method in two stages. We first perform SFT on the curated reasoning corpus to initialize the model with the desired inference protocol, so that it learns to produce structured multi-view reasoning under teacher forcing. In our implementation, we use a learning rate of $1\times 10^{-5}$ for SFT. Starting from this initialization, we then apply RL with GRPO to further optimize the policy with sampled rollouts, which better matches test-time generation and directly reinforces behaviors that lead to correct predictions. We use a smaller learning rate of $1\times 10^{-6}$ for RL to ensure stable updates, and sample $K{=}8$ trajectories per input instance to compute group-relative advantages. This two-stage pipeline combines the stability and controllability of imitation learning with the flexibility of RL to improve robustness under multi-view inputs.

\subsubsection{Reward Design for RL}
\label{sec:reward_design}

A key challenge in RL for reasoning models is that optimizing only for task correctness can induce degenerate behaviors. In multi-view settings, correctness-only training may encourage shortcut reasoning (e.g., answering from a single salient view without establishing cross-view consistency) and can also lead to unstable generations that omit required sections or collapse to overly terse outputs. To address these issues, we design a composite reward that explicitly enforces (i) answer correctness, (ii) format compliance with the intended two-stage protocol, and (iii) a reasonable response length that balances sufficient reasoning with verbosity control.

\paragraph{Answer correctness reward.}
Since all training instances are multiple-choice questions, we extract the predicted option from the \texttt{<answer>} field and use a binary reward:
\begin{equation}
\label{eq:reward_ans}
r_{\mathrm{ans}}(x,y) =
\mathbb{I}\!\left[\mathrm{ans}(y)=\mathrm{gt}(x)\right],
\end{equation}
where $\mathrm{ans}(y)$ denotes the extracted option and $\mathrm{gt}(x)$ is the ground-truth label.

\paragraph{Format validity reward.}
To stabilize multi-stage generation during RL, we enforce a strict output structure. A response is considered valid only if it contains three tag-delimited sections in the fixed order \texttt{<spatial\_thinking>}, \texttt{<thinking>}, and \texttt{<answer>}, with each tag properly opened and closed. We implement the format reward as a binary indicator:
\begin{equation}
\label{eq:reward_fmt}
r_{\mathrm{fmt}}(x,y) =
\mathbb{I}\!\left[\mathrm{ValidFormat}(y)\right],
\end{equation}
where $\mathrm{ValidFormat}(y)$ returns true if and only if all required tag pairs are present and their order is strictly consistent, without missing tags, swapped order, duplicated sections, or malformed closures. This constraint discourages format violations and prevents the policy from bypassing the intended two-stage reasoning behavior by directly emitting an answer.

\paragraph{Length regularization reward.}
Finally, we add a length-based shaping term to discourage both under-generation (insufficient reasoning content) and over-generation (unnecessary verbosity). Let $\ell(y)$ denote the length of the generated response (measured in tokens). We define a length reward that is activated only when the prediction is correct and the response length falls within a preferred interval:
\begin{equation}
\label{eq:reward_len}
r_{\mathrm{len}}(x,y)=
\begin{cases}
\omega, & \text{if } r_{\mathrm{ans}}(x,y)=1 \ \text{and}\  \ell_{\min}\le \ell(y)\le \ell_{\max},\\
0, & \text{otherwise},
\end{cases}
\end{equation}
where $\omega>0$ is the shaping weight and $[\ell_{\min},\ell_{\max}]$ specifies the target length range. In all experiments, we set $\omega=0.2$, $\ell_{\min}=320$, and $\ell_{\max}=512$.

\paragraph{Composite reward.}
We combine the above components into a single reward used by GRPO:
\begin{equation}
\label{eq:reward_total}
r(x,y) \;=\; r_{\mathrm{ans}}(x,y) \;+\; \lambda r_{\mathrm{fmt}}(x,y) \;+\; r_{\mathrm{len}}(x,y),
\end{equation}
where $\lambda \in [0,1]$ controls the strength of format regularization. This composite reward encourages correct answers with disciplined two-stage structure, while maintaining a reasonable reasoning length that avoids both premature truncation and overlong generations.

 \section{Experiments}
\label{sec:experiments}

\subsection{Experimental Setup}
\label{sec:exp_setup}

\paragraph{Implementation Details} We evaluate our method on multi-view spatial reasoning benchmarks under a unified training and inference setup. Our training follows the two-stage pipeline described earlier. In the SFT stage, we fine-tune the model using a learning rate of $1\times10^{-5}$. In the RL stage, we apply GRPO with a smaller learning rate of $1\times10^{-6}$ for stable policy updates. For each training instance in RL, we sample a group of $K{=}8$ trajectories to compute group-relative advantages. Unless otherwise specified, RL training is conducted for 1500 optimization steps.


\paragraph{Evaluation Settings.}
We evaluate our model on three multi-image, multi-view benchmarks: MMSI-Bench~\citep{yang2025mmsi}, MindCube~\citep{yin2025spatial}, and ViewSpatial-Bench~\citep{li2025viewspatial}. MMSI-Bench focuses on multi-image spatial reasoning that requires aligning evidence across views, such as viewpoint transformations and occlusion-aware inference. MindCube tests whether models can build a consistent mental model from limited views and perform perspective-sensitive reasoning under partial observability. ViewSpatial-Bench emphasizes viewpoint-dependent spatial localization and cross-view reference frames, highlighting generalization challenges under camera viewpoint shifts. All questions in these benchmarks are multiple-choice, and we therefore report accuracy as the primary metric. Unless otherwise stated, our decoding and inference hyperparameters follow the recommended settings of Qwen3-VL-4B~\citep{bai2025qwen3vltechnicalreport}.

\subsection{Quantitative Results}
\label{sec:quant_results}

\begin{table}[t]
\centering
\small
\setlength{\tabcolsep}{6pt}
\begin{tabularx}{1\linewidth}{l c Y Y Y Y}
\toprule
\textbf{Model} & \textbf{Size} & \textbf{MMSI} & \textbf{MindCube} & \textbf{ViewSpatial} & \textbf{Avg.} \\
\midrule
RandomChoice & -- & 25.0 & 33.0 & 26.3 & 28.1 \\
\specialrule{\lightrulewidth}{\aboverulesep}{0pt}

\rowcolor{black!5}\multicolumn{6}{l}{\rule{0pt}{2.3ex}\textit{Close-source models}}\\
Gemini-2.5-Pro & -- & 38.0 & 57.6 & 46.0 & 47.2 \\
GPT-5 & -- & 41.8 & 56.3 & 45.5 & 47.9 \\
Gemini-3-Pro-Preview & -- & 45.2 & 70.8 & 50.3 & 55.4 \\
\specialrule{\lightrulewidth}{\aboverulesep}{0pt}

\rowcolor{black!5}\multicolumn{6}{l}{\rule{0pt}{2.3ex}\textit{Open-source General Models}}\\
InternVL3-2B~\citep{zhu2025internvl3}& 2B & 26.5 & 37.5 & 32.5 & 32.2 \\
InternVL3-8B~\citep{zhu2025internvl3} & 8B & 28.0 & 41.5 & 38.6 & 36.0 \\
Qwen2.5-VL-3B-Instruct~\citep{bai2025qwen2} & 3B & 28.6 & 37.6 & 31.9 & 32.7 \\
Qwen2.5-VL-7B-Instruct~\citep{bai2025qwen2} & 7B & 26.8 & 36.0 & 36.8 & 33.2 \\
Qwen3-VL-2B-Instruct~\citep{bai2025qwen3vltechnicalreport} & 2B & 28.9 & 34.5 & 36.9 & 33.4 \\
Qwen3-VL-4B-Instruct~\citep{bai2025qwen3vltechnicalreport} & 4B & 30.1 & 37.0 & 42.5 & 36.5 \\
Qwen3-VL-8B-Instruct~\citep{bai2025qwen3vltechnicalreport} & 8B & 31.1 & 29.4 & 42.2 & 34.2 \\

\specialrule{\lightrulewidth}{\aboverulesep}{0pt}

\rowcolor{black!5}\multicolumn{6}{l}{\rule{0pt}{2.3ex}\textit{Spatial Intelligence Models}}\\
SpatialLadder-3B~\citep{li2025spatialladder} & 3B & 27.4 & 43.4 & 39.8 & 36.9 \\
Spatial-MLLM-4B~\citep{wu2025spatial} & 4B & 26.1 & 33.4 & 34.6 & 31.4 \\
SpaceR-7B~\citep{ouyang2025spacer} & 7B & 27.4 & 37.9 & 35.8 & 33.7 \\
ViLaSR-7B~\citep{wu2025reinforcing} & 7B & 30.2 & 35.1 & 35.7 & 33.7 \\
Cambrian-S-3B~\citep{yang2025cambrian} & 3B & 25.2 & 32.5 & 39.0 & 32.2 \\
Cambrian-S-7B~\citep{yang2025cambrian} & 7B & 25.8 & 39.6 & 40.9 & 35.4 \\
VST-3B-RL~\citep{yang2025visual} & 3B & 32.0 & 36.4 & 45.0 & 37.8 \\
VST-7B-RL~\citep{yang2025visual} & 7B & 34.8 & 39.1 & 42.4 & 38.8 \\

\specialrule{\lightrulewidth}{\aboverulesep}{0pt}

\rowcolor{blue!2}\multicolumn{6}{l}{\rule{0pt}{2.3ex}\textit{Ours}}\\
\rowcolor{blue!5} \method (SFT) \rule{0pt}{2.3ex} & 4B & 32.4 & 68.5 & 45.1 & 48.7 \\
\rowcolor{blue!10} \method (SFT + RL)\rule{0pt}{2.3ex} & 4B & \textbf{35.4} & \textbf{77.0} & \textbf{45.4} & \textbf{52.6} \\

\specialrule{\heavyrulewidth}{0pt}{\belowrulesep}
\end{tabularx}
\caption{Overall accuracy (\%) on three multi-view spatial reasoning benchmarks (MMSI-Bench, MindCube, and ViewSpatial) for a range of proprietary and open-source MLLMs, including our \method variants}
\label{tab:main_results}
\end{table}

\Cref{tab:main_results} summarizes the main results on three multi-view benchmarks. Our \method achieves the best overall performance among open-source 4B-scale models, with clear gains over the Qwen3-VL-4B family. In particular, \method (SFT+RL) improves Qwen3-VL-4B-Instruct by +5.3\% on MMSI-Bench (35.4\% vs.\ 30.1\%), and yields a large improvement on MindCube (77.0\% vs.\ 37.0\%), indicating substantially stronger cross-view reasoning that benefits from our two-stage training and GRPO alignment. On ViewSpatial, \method remains competitive at 45.4\% and consistently outperforms Qwen3-VL-4B-Instruct (42.5\%). Overall, these results demonstrate that explicitly optimizing for multi-view reasoning yields consistent gains across diverse evaluation settings.

To better understand the source of the improvements, \Cref{tab:finegrained_mmsi} reports a fine-grained breakdown on MMSI-Bench comparing \method{} with Qwen3-VL-4B-Instruct and Qwen3-VL-4B-Thinking. Overall, \method achieves a 17.6\% relative improvement over Qwen3-VL-4B-Instruct on MMSI-Bench (35.4\% vs.\ 30.1\%), suggesting that our explicit cross-view pre-alignment yields more effective multi-view reasoning rather than relying on view-local shortcuts. Notably, \method also outperforms Qwen3-VL-4B-Thinking (35.4\% vs.\ 29.0\%), a reasoning-focused model trained with large amounts of high-quality chain-of-thought data. This comparison highlights that simply encouraging longer or higher-quality deliberation is insufficient for multi-view spatial reasoning; explicitly enforcing cross-view spatial consistency provides additional and complementary benefits.

\subsection{Qualitative Analysis}
\label{sec:qualitative}
 \begin{figure}[t!]
    \centering
 
        \includegraphics[width=1\linewidth]{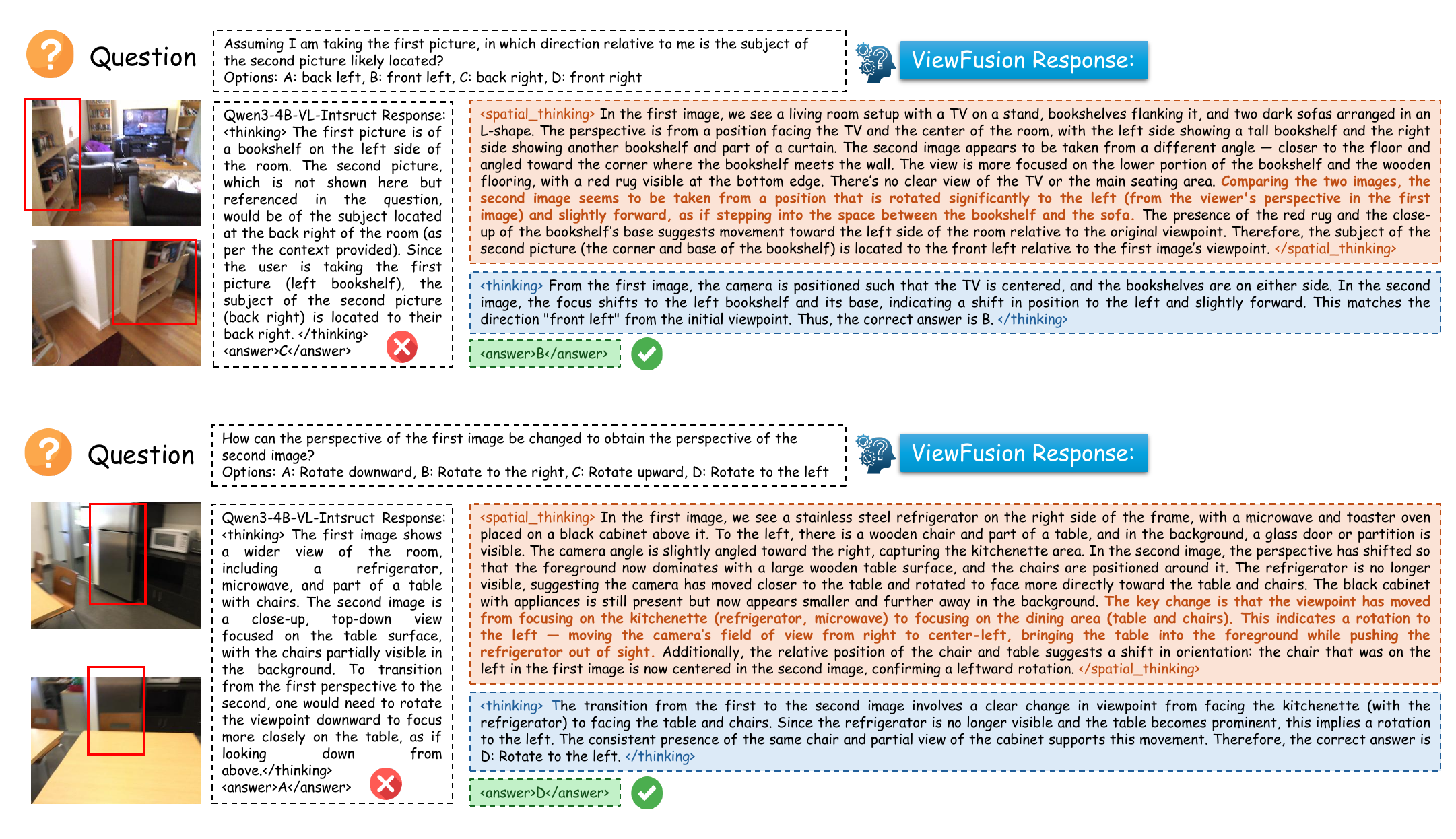}

        \caption{Qualitative examples on MMSI-Bench. The red boxes highlight the same visual elements observed from different viewpoints across the two images. Compared with Qwen3-VL-4B-Instruct,\method better aligns cross-view correspondences and infers the underlying viewpoint change, leading to correct answers.}

        \label{fig:case}
\end{figure}
Figure~\ref{fig:case} presents representative MMSI-Bench examples to illustrate why multi-view spatial reasoning requires more than view-local image description. The red boxes mark corresponding visual elements appearing across two images from different viewpoints. While a strong baseline (Qwen3-VL-4B-Instruct) can often describe salient objects within each view, it frequently fails to establish \emph{cross-view spatial consistency}---e.g., how the camera moves between views, which objects correspond under viewpoint change, and how visibility/occlusion evolves, and may therefore jump to an answer based on incomplete or mismatched evidence. In contrast, \method explicitly performs spatial pre-thinking before question solving. In the figure, the brown \texttt{<spatial\_thinking>} segment demonstrates this behavior: it links the boxed elements across views and infers the underlying viewpoint relationship (highlighted by the bold brown phrases), such as the relative rotation/translation implied by changes in object position and visibility. This intermediate cross-view alignment provides a consistent spatial interpretation that the subsequent reasoning stage can reliably condition on, leading to correct answers in cases where the answer is only recoverable by reasoning about viewpoint transformations rather than relying on a single image or a purely descriptive summary.

\subsection{Ablation Study}
\label{sec:ablation}

\begin{table}[t]
\centering
\scriptsize
\setlength{\tabcolsep}{4pt}
\renewcommand{\arraystretch}{1.15}
\resizebox{\linewidth}{!}{%
\begin{tabular}{l c c c c c c c c c c c c}
\toprule
\multirow{2}{*}{\textbf{Models}} &
\multicolumn{6}{c}{\textbf{Positional Relationship}} &
\multicolumn{2}{c}{\textbf{Attribute}} &
\multicolumn{2}{c}{\textbf{Motion}} &
\textbf{MSR} &
\textbf{Avg.} \\
\cmidrule(lr){2-7}\cmidrule(lr){8-9}\cmidrule(lr){10-11}
& Cam.--Cam. & Obj.--Obj. & Reg.--Reg. & Cam.--Obj. & Obj.--Reg. & Cam.--Reg.
& Meas. & Appr.
& Cam. & Obj.
& -- & \\
\midrule

Qwen3-4B-VL-Thinking &25.8 &26.6 &34.5 &33.7 &25.9 &36.1 &48.4 &28.8 &21.6 &26.3 &23.2 &29.0 \\
Qwen3-4B-VL-Instruct &30.1 &34.0 &29.6 &34.9 &29.4 &39.8 &45.3 &19.7 &21.6 &23.7 &26.7 &30.1 \\
\method              &46.2 &41.5 &30.9 &44.2 &21.2 &53.0 &35.9 &34.9 &40.5 &32.9 &23.2 &35.4 \\

\bottomrule
\end{tabular}
}
\caption{Fine-grained accuracy (\%) breakdown on MMSI-Bench, comparing \method with Qwen3-VL-4B-Instruct and Qwen3-VL-4B-Thinking across subcategories of positional relationships, attributes, motion, and MSR.}

\label{tab:finegrained_mmsi}
\end{table}

We conduct ablation studies on MMSI-Bench to isolate the contributions of key components as shown in~\Cref{tab:ablation}. First, replacing our structured two-stage output with free-form reasoning under RL (``Free format Reasoning + RL'') reduces overall accuracy from 35.4 to 33.4, indicating that enforcing an explicit spatial pre-thinking stage helps mitigate shortcut behaviors and improves robustness. Second, removing GRPO (``w/o GRPO'') leads to a larger drop (35.4 $\rightarrow$ 32.4), demonstrating that RL optimization with group-relative advantages is important for improving correctness under multi-view inputs beyond SFT alone. Third, removing the format reward (``w/o Format Reward'') yields a modest decrease in MMSI accuracy (35.4 $\rightarrow$ 35.0) but noticeably changes the distribution across subcategories, consistent with the role of the format reward as a stabilizer that maintains disciplined generation and prevents bypassing the intended inference protocol. Taken together, the ablations confirm that both the two-stage reasoning supervision and GRPO-based RL are necessary to achieve the strongest and most reliable multi-view spatial reasoning performance.

\begin{table}[t]
\centering
\scriptsize
\setlength{\tabcolsep}{4pt}
\renewcommand{\arraystretch}{1.15}
\resizebox{\linewidth}{!}{%
\begin{tabular}{l c c c c c c c c c c c c}
\toprule
\multirow{2}{*}{\textbf{Models}} &
\multicolumn{6}{c}{\textbf{Positional Relationship}} &
\multicolumn{2}{c}{\textbf{Attribute}} &
\multicolumn{2}{c}{\textbf{Motion}} &
\textbf{MSR} &
\textbf{Avg.} \\
\cmidrule(lr){2-7}\cmidrule(lr){8-9}\cmidrule(lr){10-11}
& Cam.--Cam. & Obj.--Obj. & Reg.--Reg. & Cam.--Obj. & Obj.--Reg. & Cam.--Reg.
& Meas. & Appr.
& Cam. & Obj.
& -- & \\
\midrule
Full Method          &46.2 &41.5 &30.9 &44.2 &21.2 &53.0 &35.9 &34.9 &40.5 &32.9 &23.2 &35.4 \\
Free format Reasoning + RL &37.6 &31.9 &25.9 &39.5 &28.2 &49.4 &46.9 &27.3 &31.1 &29.0 &28.3 &33.4 \\
w/o GRPO             &28.0 &27.7 &35.8 &37.2 &29.4 &47.0 &45.3 &30.3 &32.4 &25.0 &27.8 &32.4 \\
w/o Format Reward    &40.9 &30.9 &33.3 &46.5 &32.9 &51.8 &29.7 &40.9 &37.8 &34.2 &22.7 &35.0 \\
\bottomrule
\end{tabular}
}
\caption{Ablation study on MMSI-Bench (accuracy, \%), analyzing the impact of free-form reasoning, removing GRPO, and removing the format reward on fine-grained subcategories and overall performance.}

\label{tab:ablation}
\end{table}

\subsection{Training Curves}
\label{sec:training_curves}

\begin{figure}[t]
    \centering
    \begin{minipage}[t]{0.24\linewidth}
        \centering
        \includegraphics[width=\linewidth]{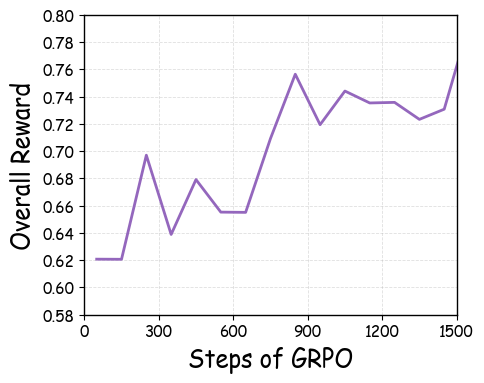}
    \end{minipage}\hfill
    \begin{minipage}[t]{0.24\linewidth}
        \centering
        \includegraphics[width=\linewidth]{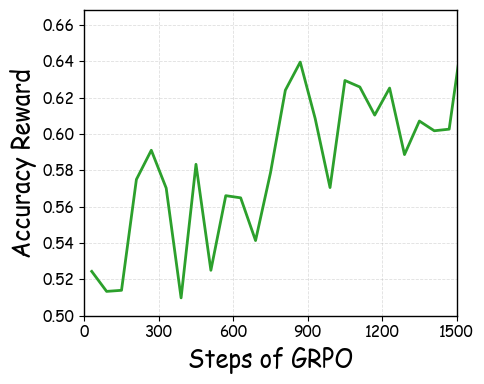}
    \end{minipage}\hfill
    \begin{minipage}[t]{0.24\linewidth}
        \centering
        \includegraphics[width=\linewidth]{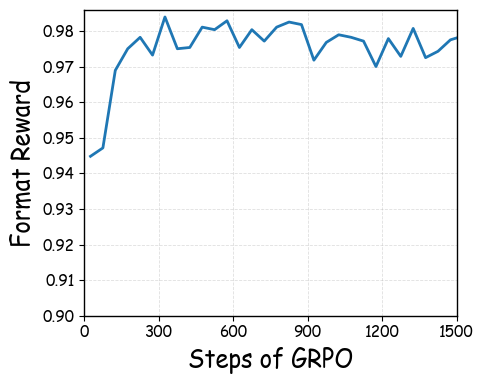}
    \end{minipage}
        \begin{minipage}[t]{0.24\linewidth}
        \centering
        \includegraphics[width=\linewidth]{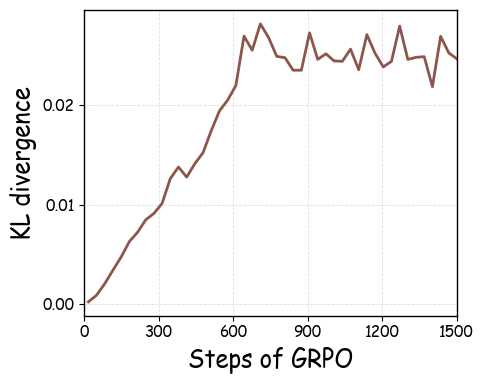}
    \end{minipage}
     \caption{Training curves during GRPO over 1500 steps, including the total reward (left), the accuracy reward (second), the format reward (third), and the KL divergence to the reference policy (right).}

    \label{fig:training_curves}
\end{figure}

\Cref{fig:training_curves} shows the optimization dynamics of the GRPO stage over 1500 steps. The total reward increases steadily, indicating that the policy is progressively improving under the combined objective. Decomposing the reward reveals that the accuracy reward exhibits a clear upward trend, while the format reward quickly reaches and maintains a high level, suggesting that the desired output discipline is learned early and remains stable throughout RL training. Meanwhile, the KL divergence increases during the initial phase and then plateaus, implying that the policy explores beyond the reference model but remains controlled under KL regularization. Overall, these curves demonstrate stable RL training that improves correctness while preserving the intended structured behavior.
 \section{Conclusion}
\label{sec:conclusion}

We presented \method, a two-stage ``think twice'' framework for multi-view spatial reasoning that makes cross-view alignment an explicit first step rather than an implicit byproduct of question answering. By synthesizing structured supervision for spatial pre-thinking and further optimizing with GRPO using a correctness reward and a strict format reward, our approach mitigates shortcut behaviors that underuse available viewpoints and stabilizes multi-stage generation. Experiments on three multi-view benchmarks demonstrate consistent improvements, with particularly strong gains on MMSI-Bench and MindCube, and fine-grained analyses show that the benefits are concentrated in categories that require genuine viewpoint reasoning. Ablations and qualitative examples further validate the contribution of each component and highlight that improved deliberation alone is insufficient without explicit cross-view spatial consistency. We hope \method serves as a simple, practical step toward more reliable multi-view reasoning in MLLMs, and motivates future work on scalable cross-view alignment objectives and broader spatial generalization.


\newpage
\bibliography{iclr2026_conference}
\bibliographystyle{colm2025_conference}

\newpage
\appendix
\renewcommand \thepart{} 
    \renewcommand \partname{}
\part{Appendix} 

\section{Prompt Template}
\label{sec:appendix_prompt}

We use the following prompt template for multi-view spatial reasoning:

\begin{tcolorbox}[
  colback=black!5,
  colframe=black!20,
  boxrule=0.4pt,
  arc=2pt,
  left=6pt,right=6pt,top=6pt,bottom=6pt
]
\small
\texttt{\{Question\}}

\vspace{0.6em}
Please reason step by step. First, you need to provide a description of the perspective transformation applied to this set of images. Briefly explain how the perspective changes between the images (e.g., rotated 90 degrees clockwise), and describe the relative positions of the objects based on the background and occlusion relationships. Provide your description process and enclose it within the \texttt{<spatial\_thinking>} \texttt{</spatial\_thinking>} tags.

\vspace{0.6em}
When the perspective changes, use multiple images to discover objects that are not visible in a single image, and provide your detailed reasoning for the raw question within \texttt{<thinking>} \texttt{</thinking>}.

\vspace{0.6em}
Provide only the single option letter (e.g., A, B, C, D) within the \texttt{<answer>} \texttt{</answer>} tags
\end{tcolorbox}
    \parttoc 
\end{document}